\documentclass[a4paper, 11pt]{article} 

\usepackage[utf8]{inputenc}
\usepackage[numbers]{natbib}
\usepackage[spanish]{babel}
\usepackage[protrusion=true,expansion=true]{microtype} 
\usepackage{graphicx} 
\usepackage{wrapfig} 

\usepackage{subfigure}
\usepackage{graphicx}      
\usepackage{natbib}        
\usepackage{amsmath,amssymb,amsthm}
\usepackage[dvipsnames]{xcolor}
\usepackage{pstricks}
\usepackage{enumerate}
\usepackage{arydshln}
\usepackage{tikz}
\usepackage{array}
\usepackage{multirow}
\usepackage{lmodern,textcomp}

\makeatletter
\renewcommand{\@listI}{\itemsep=0pt} 

\renewcommand{\maketitle}{ 
	\begin{flushright} 
		{\LARGE\@title} 
		
		\vspace{50pt} 
		
		{\large\@author} 
		\\\@date 
		
		\vspace{40pt} 
	\end{flushright}
}

\title{\large{\textbf{Diseña, Fabrica y Programa Tu Propio Robot}}\\ 
	Propuesta  y guía de recomendaciones} 

\author{\textsc{Leopoldo Armesto} 
	\\{\textit{Universitat Politènica de València}}} 

\date{\today} 
\begin{document}

\maketitle

\begin{abstract}
	En este documento os presentamos a DYOR, un robot con propósitos educacionales que podréis hacer en casa o en el aula. El robot surge como consecuencia de actividades realizadas por alumnos míos en sus proyectos Final de Grado y Tesinas de Máster y dirigidas por mí y que han sido implantadas en asignaturas que imparto en la Universidad, en la que los alumnos diseñan, fabrican y programan su propio robot con materiales de bajo coste.
\end{abstract}

\hspace*{3,6mm}\textit{Keywords:} \small Robot , Arduino , Impresión 3D , Fabricación Digital , Electrónica, STEM, Programación por Bloques 

\vspace{30pt} 

\section{Introducción}

La cultura del Open Source o software libre, que permite el intercambio global de la información, se ha extendido a nuevas áreas en la última década. Por un lado, encontramos lo que ya se denomina \emph{Open Hardware} \cite{1341390}, que consiste en el acceso libre a planos de construcción de objetos físicos para ser reproducidos y mejorados. Uno de los principales caso de éxito que se ha dado en el ámbito de la electrónica de microcontroladores es el sistema Arduino \cite{Arduino} que permite de una manera sencilla y económica desarrollar proyectos que requieran interacción con el medio físico. En este ámbito podemos encontrar también el proyecto
RepRap \cite{bath18662} cuyo objetivo es el de crear máquinas que sean capaces de replicarse a sí mismas y que ha contribuido significativamente al uso de las impresoras
3D \emph{de escritorio}. Este fenómeno, junto a la democratización del uso de otras tecnologías de fabricación digital (impresión 3D, corte por láser, CNC, etc...) ha sido acuñada por varios
autores como la \emph{tercera revolución industrial} \cite{Economist}.

Una de las consecuencias de la fabricación digital es que la Comunidad Robótica tiene la oportunidad de alcanzar a un público mucho más amplio. En este
sentido, la impresión 3D es una tecnología relativamente incipiente, en un estado semi-maduro, que está teniendo un impacto importante en el campo de la robótica. Actualmente, es relativamente sencillo el proceso de descargar gran variedad de modelos de robots imprimibles, también conocidos como \emph{printbots} que pueden ser utilizados en investigación y diversas actividades académicas. Estos robots son mucho más que simples juguetes y pueden ser utilizados en múltiples formas como herramientas para los estudios de ingeniería y pre-ingeniería. Pueden ser incorporados en cursos de robótica convencional y en todo tipo de proyectos educativos. Además, los \emph{printbots} permiten generar proyectos ambiciosos que pueden
ser llevados a cabo por grupos de estudiantes que, debidamente coordinados, pueden trabajar de forma complementaria en diferentes actividades multidisciplinares.

\section{Proyecto DYOR}

\begin{figure}
	\centering
	\includegraphics[width=0.6\columnwidth]{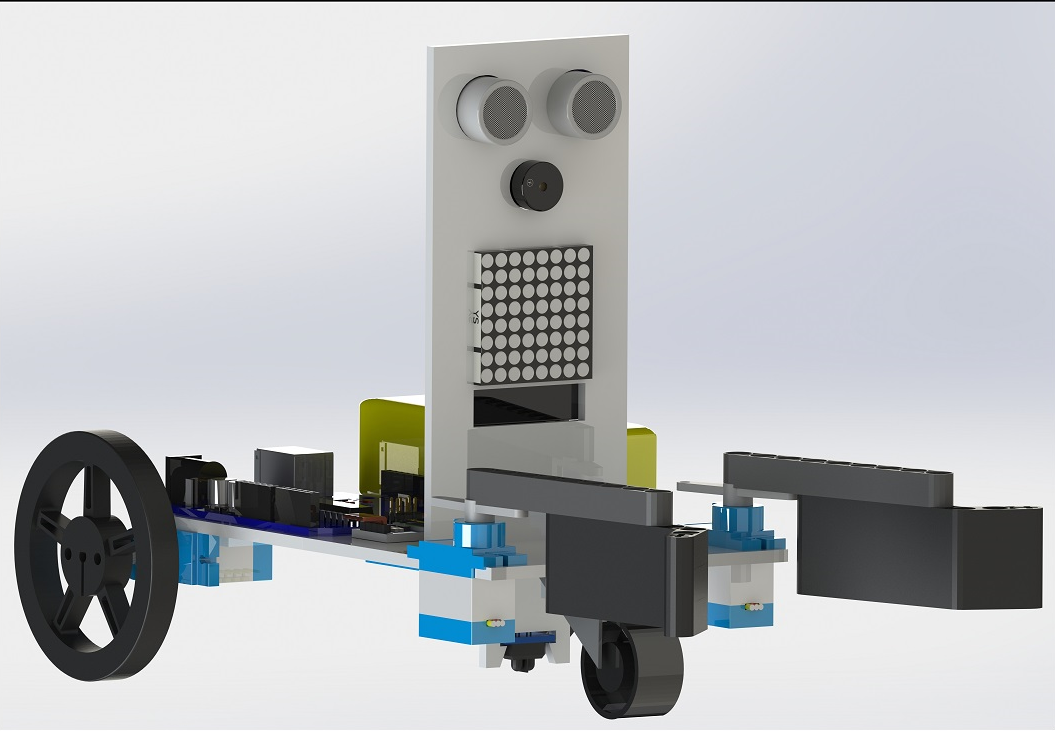}
	\caption{DYOR: Do Your Own Robot}\label{fig:dyor}
\end{figure}

\begin{figure}
	\subfigure[\textbf{C3PO}]{\includegraphics[width=0.3\columnwidth]{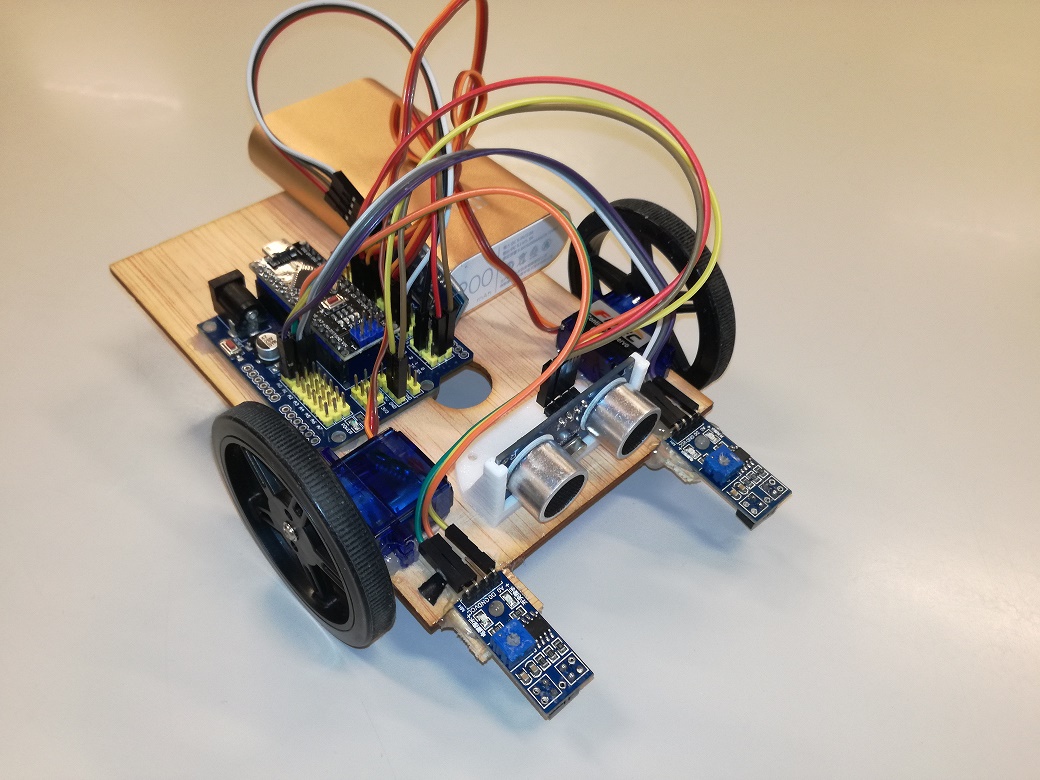}}
	\subfigure[\textbf{Rufus2}]{\includegraphics[width=0.3\columnwidth]{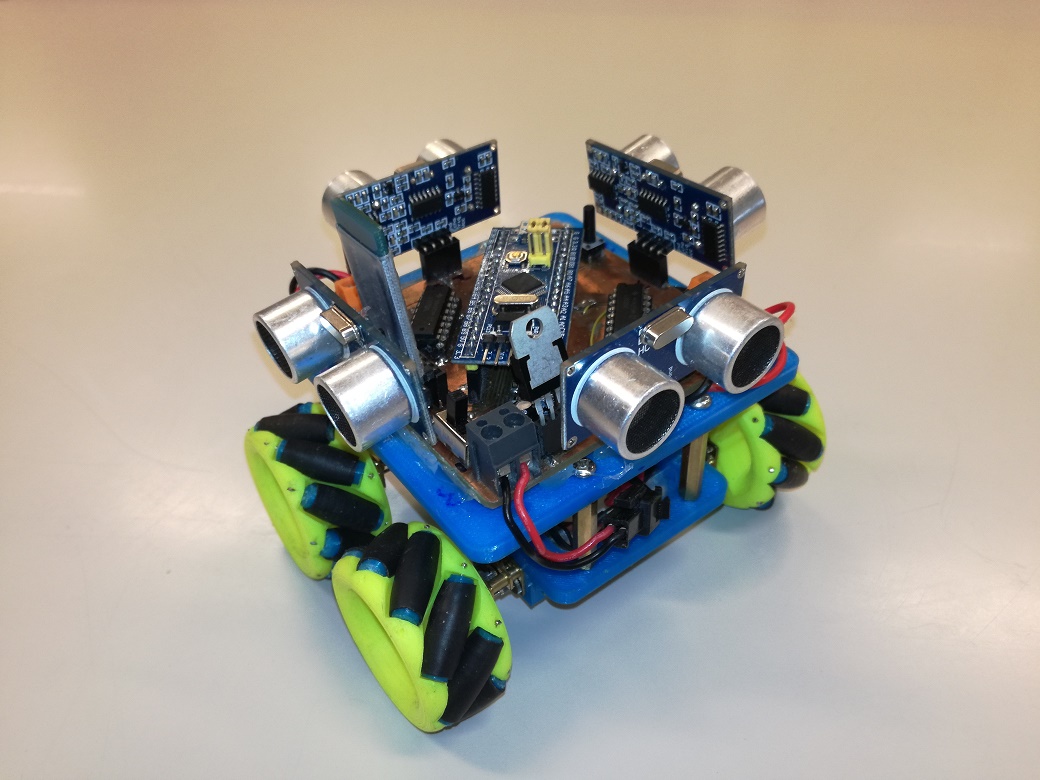}}
	\subfigure[\textbf{BeetleBot}]{\includegraphics[width=0.3\columnwidth]{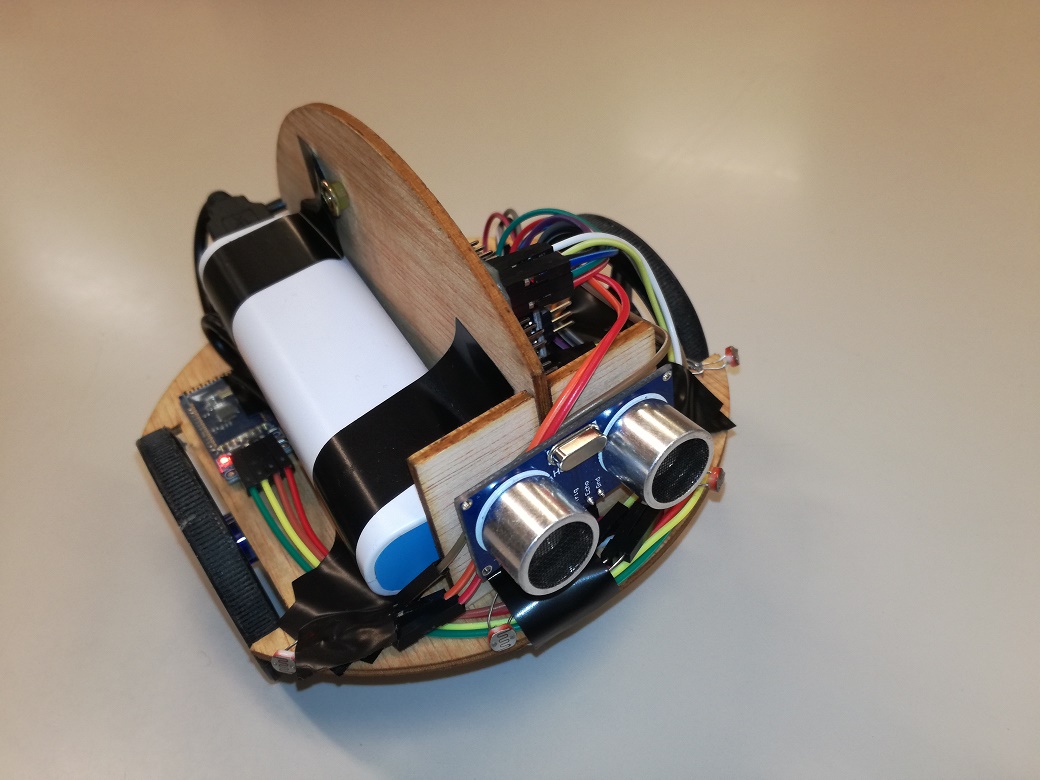}}\\
	\subfigure[\textbf{Sargal!}]{\includegraphics[width=0.3\columnwidth]{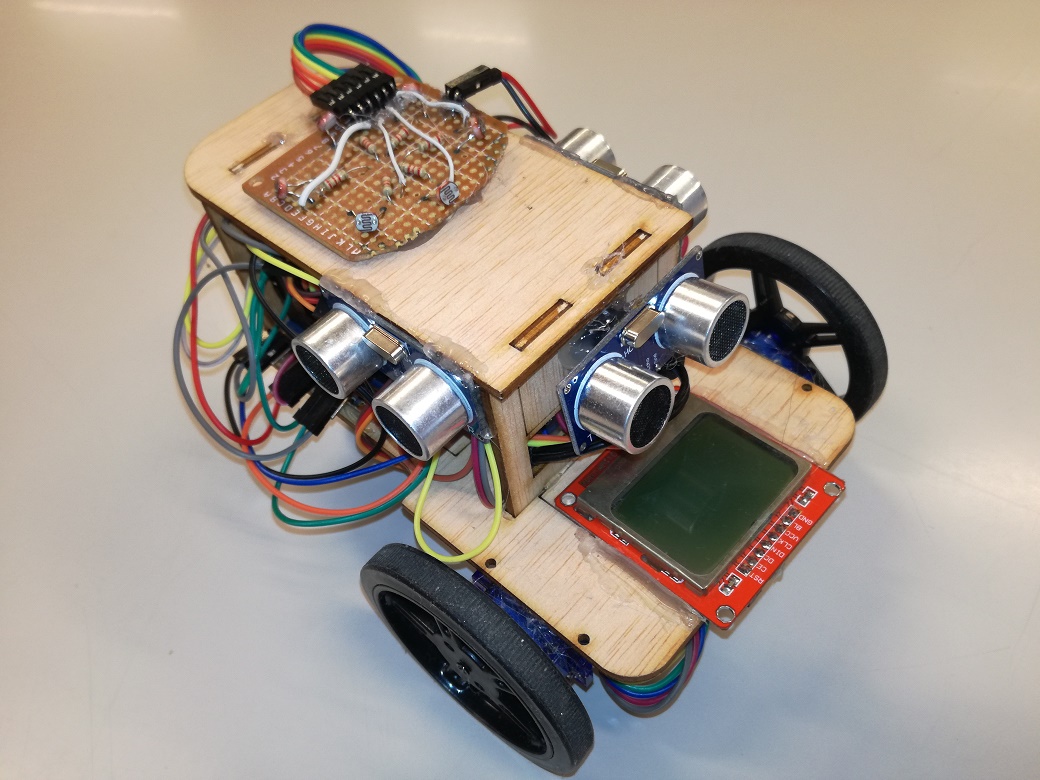}}
	\subfigure[\textbf{Jimmy}]{\includegraphics[width=0.3\columnwidth]{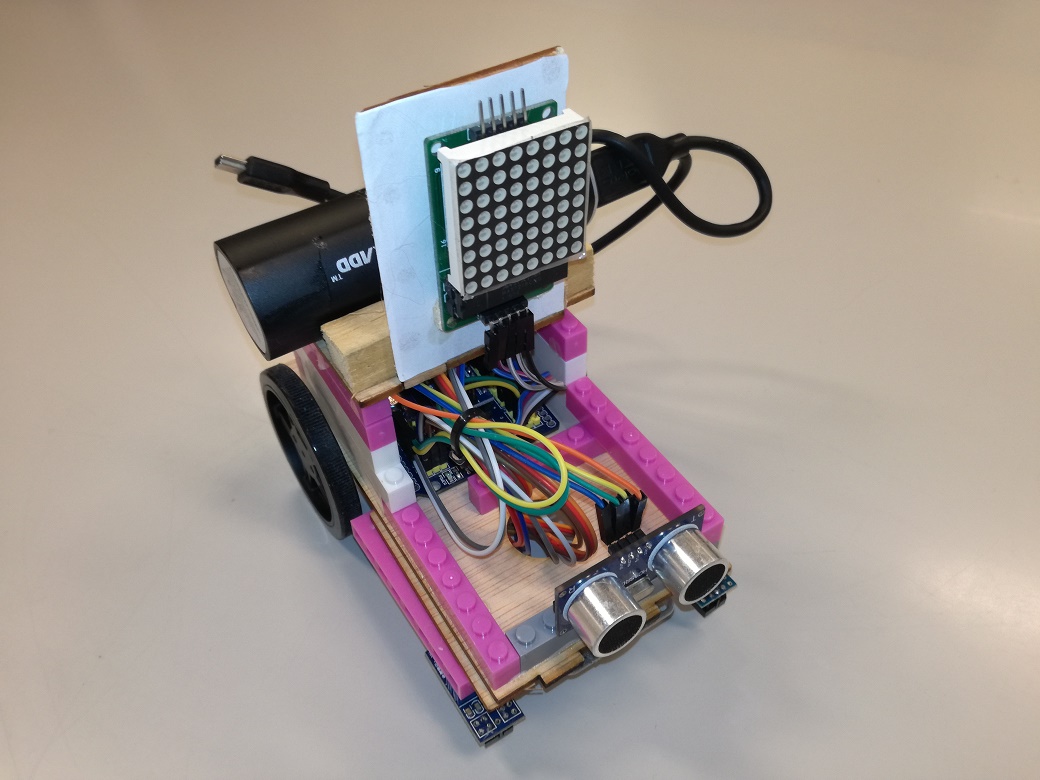}}
	\subfigure[\textbf{Wall-E-ncia}]{\includegraphics[width=0.3\columnwidth]{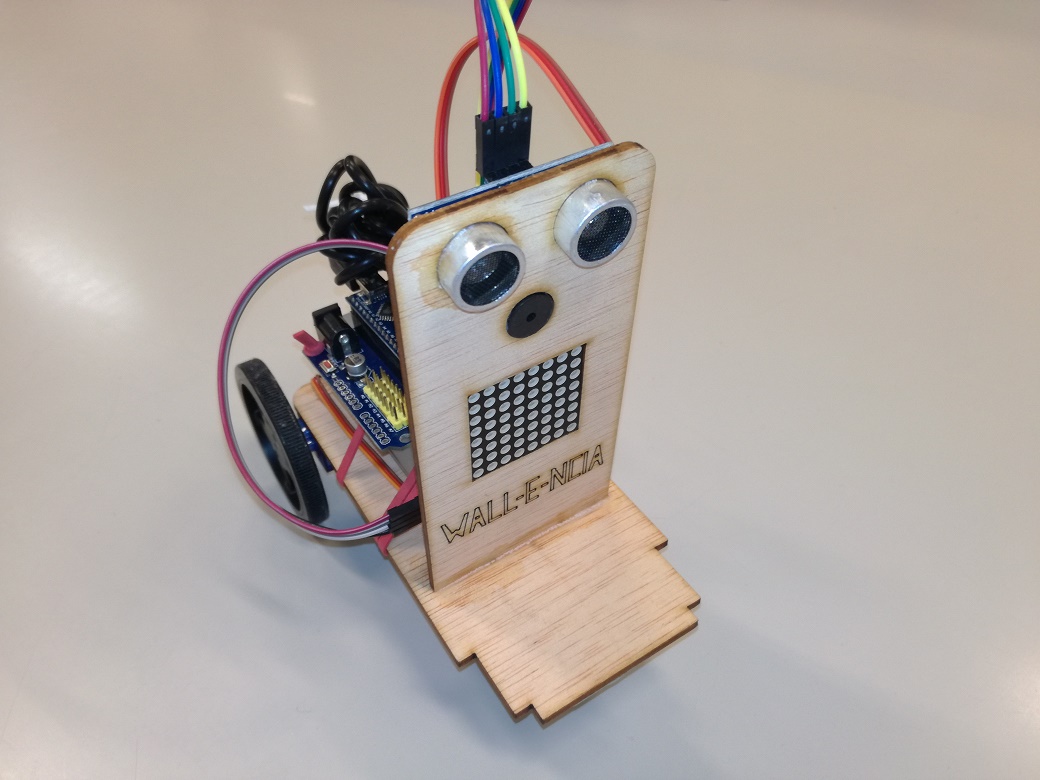}}\\
	\subfigure[\textbf{Minionbot}]{\includegraphics[width=0.3\columnwidth]{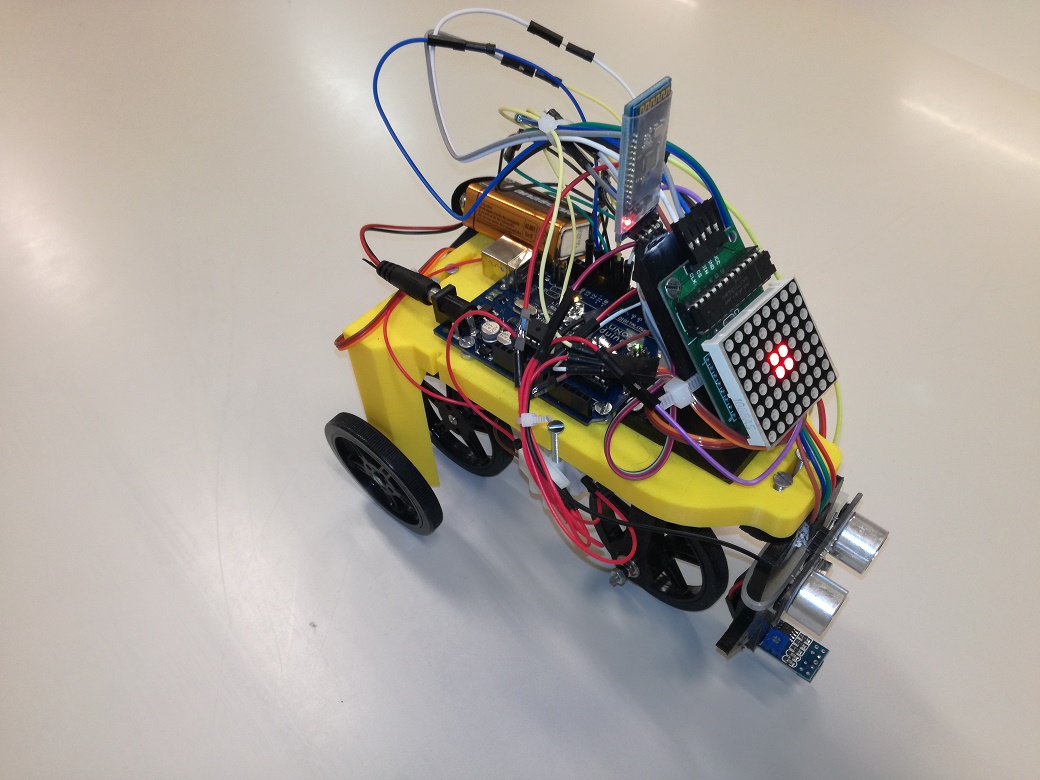}}
	\subfigure[\textbf{D-bot}]{\includegraphics[width=0.3\columnwidth]{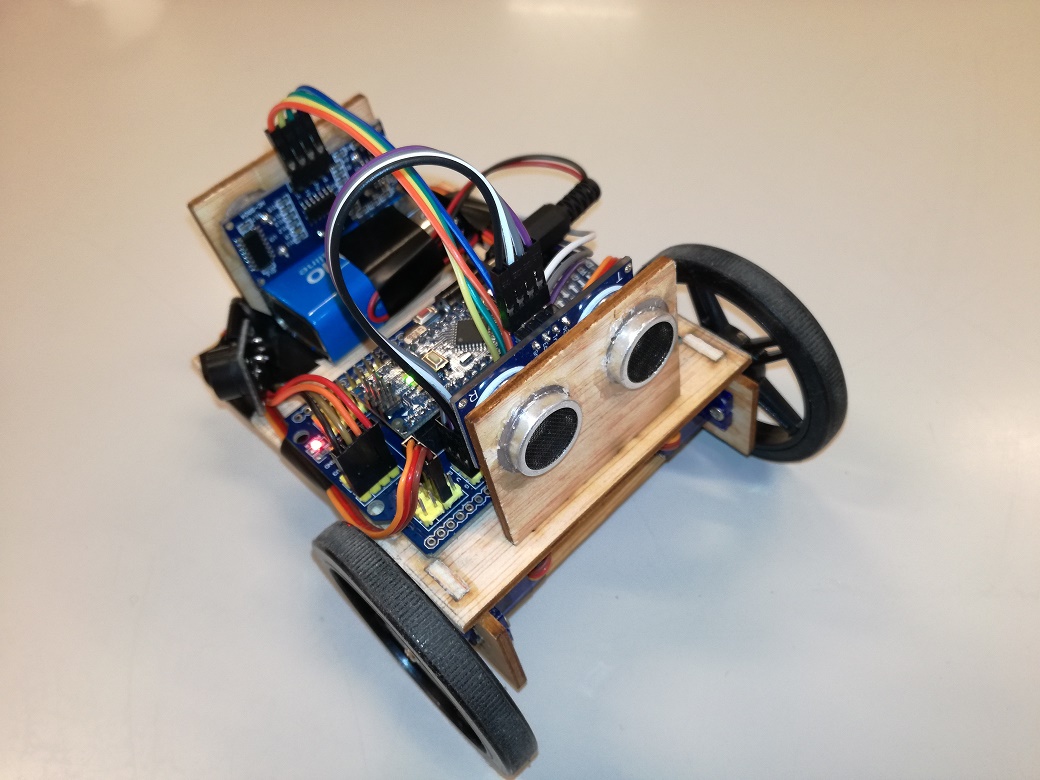}}
	\subfigure[\textbf{Blitzcranko sama}]{\includegraphics[width=0.3\columnwidth]{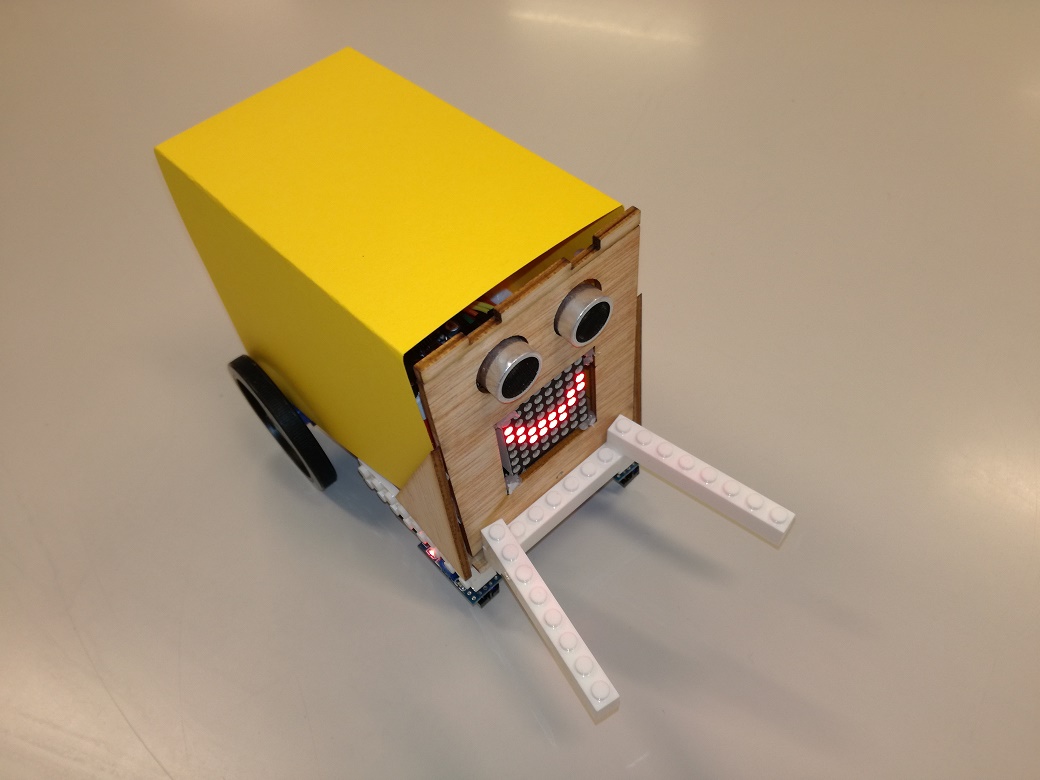}}\\
	\subfigure[\textbf{Marvizz}]{\includegraphics[width=0.3\columnwidth]{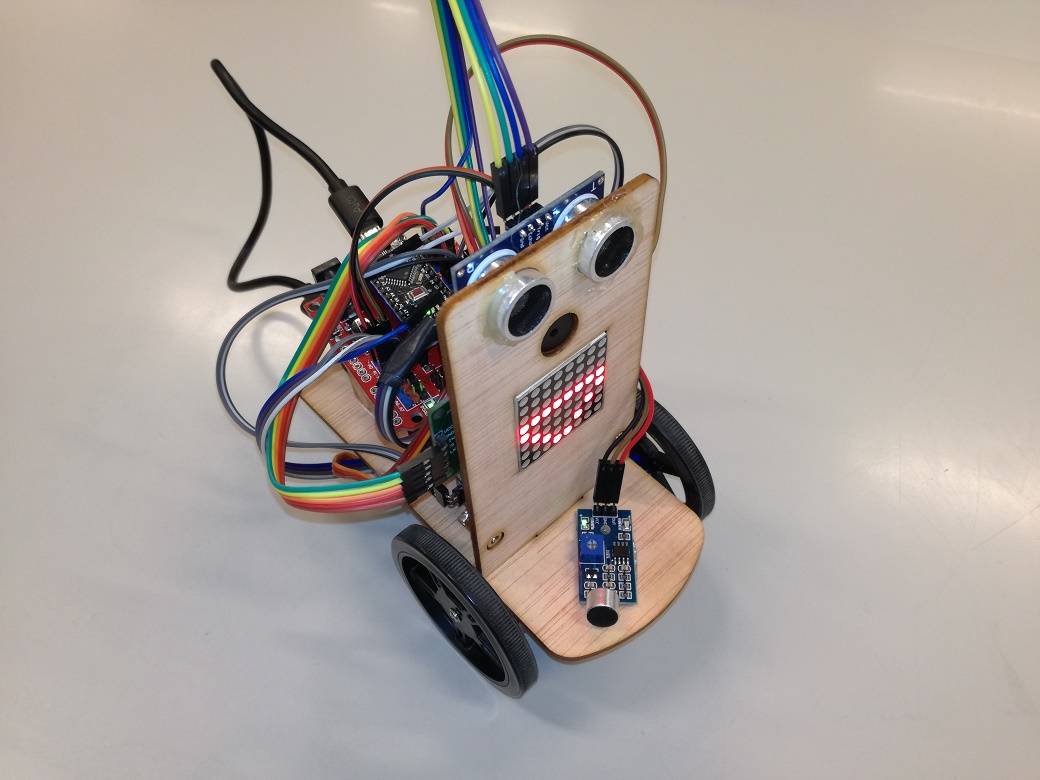}}
	\subfigure[\textbf{Robot Pirata}]{\includegraphics[width=0.3\columnwidth]{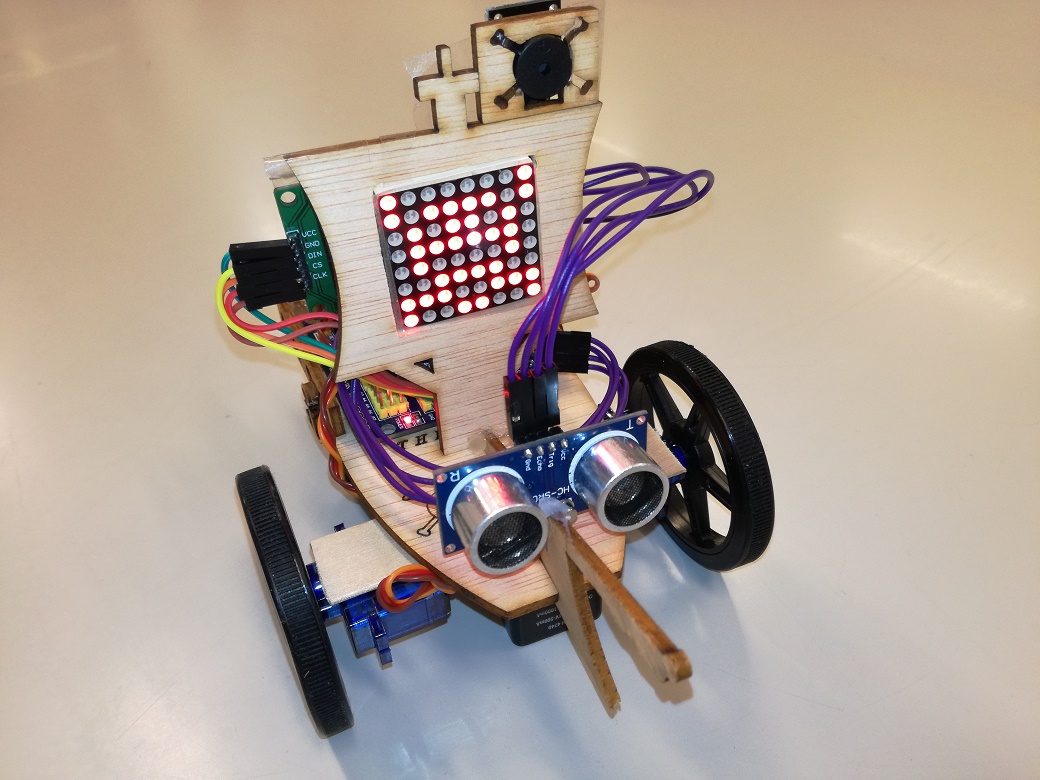}}
	\subfigure[\textbf{Illumibot}]{\includegraphics[width=0.3\columnwidth]{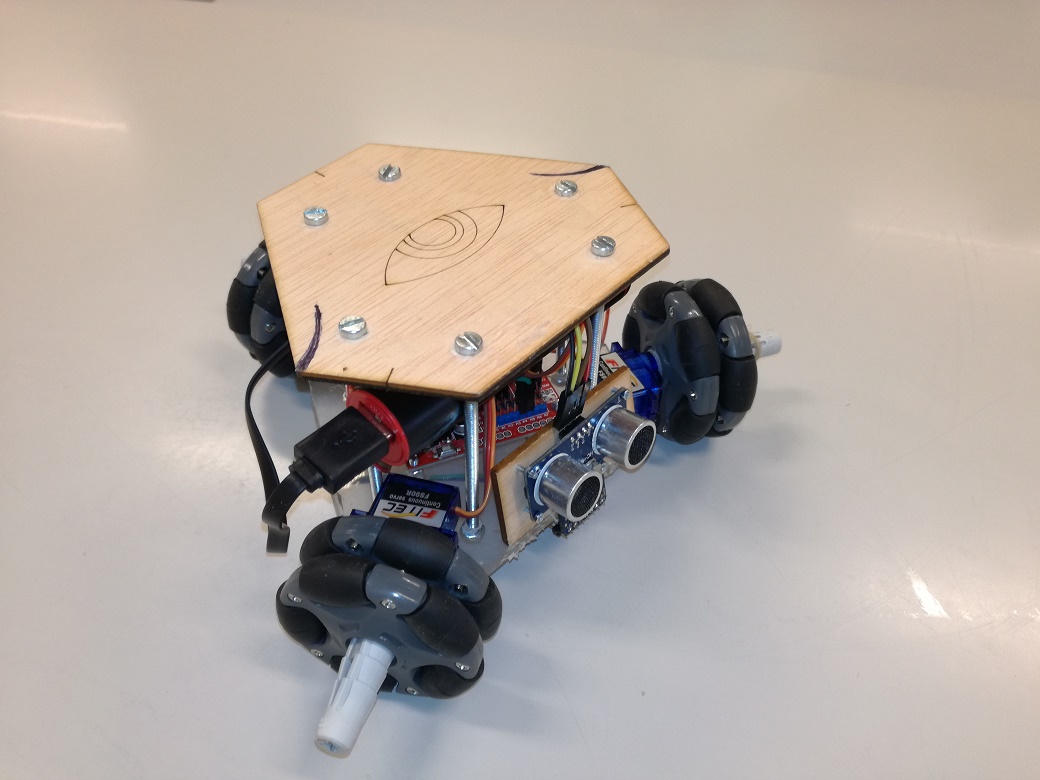}}
	\caption{Robots diseñados por alumnos míos en la Universidad utilizando la metodología propuesta (incorporando sus propias modificaciones)}\label{fig:dyor_upv}
\end{figure}

DYOR (Do Your Own Robot), ver Figura \ref{fig:dyor}, es un paquete educativo, mucho más que un simple kit de robótica. Lo he creado junto con mis alumnos en la Universitat Politècnica de València con el propósito de ser implantado como actividad dentro del currículum de Secundaria, Formación Profesional, talleres de robótica, pero también para que personas autodidactas realicen su propio robot y que aprendan las bases de la ingeniería. La Figura \ref{fig:dyor_upv} muestra un conjunto de robots diseñados por alumnos de la UPV en las asignaturas que imparto.

Asociado a DYOR, he creado este curso MOOC titulado \textbf{DYOR: Diseña, fabrica y programa tu propio robot} con el objetivo de extender este paquete educativo a todas las posibles personas interesadas. Es probablemente un curso diferente a muchos otros cursos que podréis realizar, ya que no se centra exclusivamente en una herramienta o tecnología, si no que más bien, proporciona la base de un conjunto de herramientas que os permitirán lograr los objetivos del curso: que diseñéis, fabriquéis, montéis y programéis vuestro propio robot desde cero. El curso en la actualidad es en español, pero está previsto realizar la versión equivalente en inglés en breve.

Aunque no de forma exclusiva, el curso es idóneo para profesores de centros educativos de diversa índole, cada uno adaptándolo a su manera. En España, Institutos de Educación Secundaria o IES, algunas especialidades de Formación Profesional, talleres de Robótica para niños, jóvenes o adultos, lo pueden perfectamente integrar dentro de sus currículums, adaptándolo cada uno a sus propios objetivos.

Con objeto de poder dar servicio a la posible demanda de material necesario para el curso, la empresa Robótica Fácil, dispone de kits del robot DYOR \cite{dyor_tienda} y de un portal de formación \cite{dyor}\footnote{El propósito del portal de formación es proporcionar documentación adicional, así como recopilar material de robots DYOR existentes con objeto de que os sirva de inspiración.}. Sois libres en cualquier caso de comprarles material o adquirirlo por vuestra cuenta.

Así pues, este documento ha sido redactado para describir la solución propuesta en el curso y además se proporcionan un conjunto de recomendaciones para el diseño CAD, fabricación digital, selección de componentes de electrónica y programación de un robot antes de decidir si su interés por el curso en sí mismo. Además, es importante remarcar que este documento incluye reflexiones personal de índole personal como consecuencia de haber realizado múltiples proyectos de robótica, haber trabajado con niños y estudiar tecnologías de programación por bloques que los niños utilizan como Scratch, Lego Mindstorms, BitBloq, entre otras. En cualquier caso, mi propuesta puede no coincidir con la vuestra y por tanto es normal y comprensible hacer planteamientos diferentes a los que se realizan en este curso.

Uno de los retos principales a los que nos hemos enfrentado ha sido el hecho de poder proporcionar una solución sencilla y personalizable, que utilice materiales de bajo coste basados en Arduino, que pueda re-aprovecharse para otros cursos y que al mismo tiempo permita a todo tipo de alumnos aprender las bases de la ingeniería a partir de actividades formativas que incluyen el diseño CAD, la fabricación, el montaje y la programación del robot, sin necesidad de conocimientos previos exigentes. El robot que surge de DYOR es modificable, adaptable y personalizable y se puede fabricar utilizando impresoras 3D, servicios de corte por láser o mediante el uso moldes de silicona y resinas. No obstante, al igual que mis alumnos en la Universidad, en este curso tenéis plena autonomía en las decisiones de diseño y selección de electrónica.

Quisiera remarcar que DYOR no es un robot de competición, al menos no ha sido diseñado para ello, aunque es capaz de poder hacer algunas de las típicas pruebas exigidas en las competiciones de robots, su electrónica no está optimizada para la resolución óptima de estas pruebas, si no que ha sido escogida, como ya he indicado, por su sencillez de programación y montaje.

Si sois educadores, en mi opinión, no todos vuestros alumnos serán informáticos, electrónicos o robóticos en un futuro próximo, y por tanto la actividad que he planteado ha sido diseñada para que todos puedan seguirla con facilidad, mediante el uso de herramientas pre-ingenieriles adaptadas para niños y jóvenes como TinkerCAD \cite{tinkercad}, Fritzing \cite{fritzing}, Facilino \cite{facilino} y App Inventor2 \cite{AI2}. De hecho, de forma idónea, la actividad debería ser coordinada desde varias áreas dentro del centro educativo. Así pues, desde el área de matemáticas, se podría plantear parte de los cálculos necesarios, en asignaturas que trabajan el dibujo técnico pueden claramente trabajar con el diseño CAD. En asignaturas propiamente relacionadas con la tecnología, se pueden trabajar todos los aspectos relacionados con la fabricación y la electrónica, mientras que en asignaturas de informática, se puede trabajar los contenidos de programación, robótica y diseño de Apps.

\section{Diseño CAD y Fabricación Digital}

\begin{figure}
	\centering
	\includegraphics[width=0.7\columnwidth]{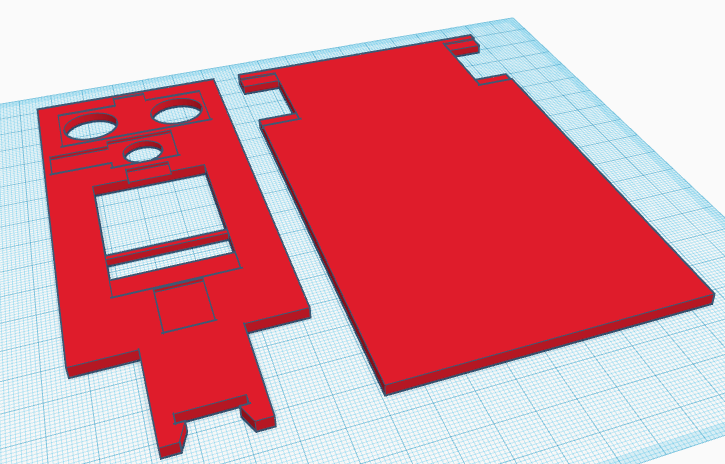}
	\caption{Diseño CAD del robot DYOR realizado con TinkerCAD}\label{fig:diseno_CAD}
\end{figure}

En el curso se plantea el diseño CAD del robot la fabricación mediante técnicas de fabricación digital: impresión 3D o corte por láser. Se os pedirá que partiendo del diseño que yo propongo o uno propio tratéis de hacer alguna modificación que os permita identificar el robot como propio. En la Figura \ref{fig:diseno_CAD} muestro la propuesta de diseño del robot. Como podéis observar el diseño del robot es mejorable en muchos aspectos, pero está hecho de forma que se pueda realizar con un conjunto de instrucciones sencillas. De hecho, observaréis a lo largo del documento que aparecen variaciones del diseño propuesto, como consecuencia de modificaciones y mejoras realizadas.

Existen varias herramientas de diseño CAD que podríais utilizar para este propósito. Tradicionalmente se trabajaba con diseño 2D, si bien, las nuevas tecnologías, parece que piden un buen dominio de diseño de objetos en 3D. Es por ello que, de entre todas las posibles opciones, lo he planteado con TinkerCAD, un programa de libre acceso que ha sido diseñado por AutoDesk, una reconocida empresa en el mundo de la ingeniería por su famoso programa AutoCAD. 

TinkerCAD es un programa de diseño CAD pensado para que lo utilicen niños y jóvenes desde edades bien tempranas, pero también lo utilizan adultos con poca experiencia en el diseño CAD. Además, dispone de una comunidad de diseñadores bastante activa, de la cual podéis extraer ideas e incorporarlas a vuestros propios proyectos. Por todo ello,  considero que TinkerCAD es un buen programa para aprender a manejar conceptos básicos del diseño CAD, ya que es intuitivo y fácil de utilizar. En el caso de que TinkerCAD sea insuficiente, su \emph{hermano mayor} es Fusión 360 (antiguamente conocido como 123D).

Otros programas como SketchUp o QCAD podrían también utilizarse para la actividad que planteo, pero me parecen menos intuitivos para su uso por niños y jóvenes y por este motivo no están incluidos en el curso. Al fin y al cabo, tampoco pretendo abarcar todos los posibles programas a utilizar.

\subsection{¿Qué restricciones de diseño tengo?}

Disponemos, por lo general, de restricciones de diseño que están principalmente condicionadas a los materiales seleccionados y a los procesos de fabricación que vayamos a utilizar. Por tanto, debemos de considerar TODOS los posibles elementos que puedan influir en nuestro diseño desde el principio. Esto es, en definitiva, un ejercicio de reflexión que se debe de hacer nada más comenzar y que debemos de tener presente en todo momento.

Por ello, recomiendo, que conozcáis bien la electrónica que vais a utilizar y sus dimensiones (buscando en Internet y en hojas de datos de los fabricantes/vendedores). También debéis de conocer la tecnología de fabricación que pensáis utilizar, ya que vuestro diseño podría estar fuertemente condicionado por esta tecnología.

Concretamente, aunque esto es una restricción \emph{suave}, sugiero un tamaño de 150x150mm${^2}$ para el diseño de vuestro robot. Los motivos son varios:
\begin{itemize}
	\item Por lo general, un mayor tamaño implica mayores costes de fabricación.
	\item Reducción del tamaño del robot, lo que implica mayor maniobrabilidad y usabilidad.
	\item En la fabricación por impresión 3D, la gran mayoría de impresoras 3D no pueden imprimir trabajos más grandes de 200x200mm${^2}$ (y es aconsejable no imprimir en los límites de la impresora), con lo que optimizar el tamaño de las piezas es fundamental para el ahorro de tiempos de fabricación. De lo contrario debemos dividir el trabajo en trabajos más sencillos lo cual implica muchas más horas de fabricación.
	\item En la fabricación por corte por láser, es fácil colocar los diseños de varios alumnos en forma tabular. Por ejemplo, en una tabla de madera de 600x900mm${^2}$\footnote{El tamaño físico de la tabla de madera deberá ser mayor porque las cortadoras láser requieren de unos márgenes mínimos.} os podrían caber hasta 24 robots.
\end{itemize}

\subsection{¿Impresión 3D o corte por láser?}
La fabricación mediante impresión 3D es idónea para la fabricación de prototipos. Tanto si estáis en clase como en casa, hoy en día no es difícil tener acceso a una impresora 3D o en su defecto se puede subcontratar un servicio de impresión 3D on-line o en centros Makers de fabricación (FabLabs).

Lo cierto es que sigue pareciendo \emph{magia}, ver que un modelo CAD que uno ha diseñado en el ordenador acaba fabricándose de forma tan sencilla con una impresora 3D (si lo comparamos a la fabricación tradicional por CNC). No obstante, debéis de ser conscientes de que la impresión 3D tiene sus propias limitaciones y os podría causar problemas si no conocéis bien la máquina o los materiales que estáis utilizando, la propia naturaleza del proceso de impresión 3D y/o el tipo de diseño (quizás no es adecuado para su impresión).

Por ejemplo, nosotros, en nuestro departamento, tenemos una impresora Da Vinci XYZ. Como muchas de las impresoras `cerradas' tienen muchas facilidades con respecto a las impresoras `abiertas', pero aún así no están exentas de problemas. Los costes de los cartuchos son significativamente más caros que los libres y además incorporan medidas de anti-pirateo que en ocasiones hacen creer a la máquina que el cartucho está finalizado, cuando en realidad no lo está. Por otro lado, las impresoras `abiertas' requieren ser un usuario con conocimientos medios/avanzados y dominar los procesos de impresión 3D, ya que de lo contrario pueden llegar a fallar con más frecuencia, por lo general, como consecuencia de una parametrización inadecuada o mantenimiento insuficiente.

Por lo general, el uso de impresoras 3D consumirá un numero de horas considerables, con lo que aquí es donde la fabricación por corte por láser se convierte en una alternativa más que interesante, ya que al final se traduce en un ahorro considerable de horas de trabajo y por tanto de costes. La principal desventaja es que condiciona vuestro diseño a 2D, pero con algo de creatividad se pueden también obtener resultados espectaculares. Por tanto, si estáis pensando en montar varios robots en el aula, incluso teniendo acceso a una impresora 3D, puede que compense subcontratar la fabricación mediante tecnología de corte por láser y utilizar la impresora 3D para validar los primeros prototipos.

De hecho, en el caso de que queráis fabricar por corte por láser,  propongo utilizar TinkerCAD y luego exportar el diseño por capas a ficheros SVG (gráfico vectorial). Soy consciente de que lo razonable en ese caso sería haber hecho el diseño directamente en 2D, pero lo he planteado así porque de esta forma hasta los más pequeños lo pueden llegar a fabricar al mismo tiempo que aprenden una herramienta de diseño 3D.

\begin{figure}
	\centering
	\subfigure[Impresión 3D]{\includegraphics[width=0.4\columnwidth]{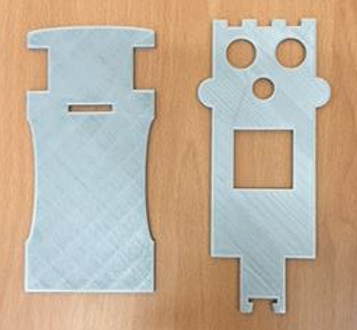}}
	\subfigure[Corte por láser fabricado en madera FDM]{\includegraphics[width=0.4\columnwidth]{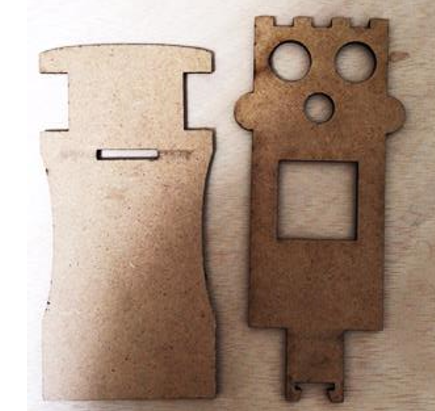}}
	\caption{Ejemplos de fabricación por impresión 3D y corte por láser.}\label{fig:fabricacion_digital}
\end{figure}

Además, tened en consideración que, por lo general, el diseño que habéis realizado para impresión 3D puede que no os valga exactamente para corte por láser, ya que el primero añade material y tiene a depositar más del que le corresponde, mientras que el segundo realiza un corte de material y acaba eliminando un poco más del que le corresponde. Lo cierto es que para las tolerancias en las que yo me manejo, este factor no suele ser relevante, pero que debería tenerse en consideración si se quieren hacer las cosas correctamente. En la Figura \ref{fig:fabricacion_digital} muestro el resultado de impresión 3D y corte por láser.

\subsection{Moldes de silicona y resinas}

\begin{figure}
	\centering
	\includegraphics[width=0.6\columnwidth]{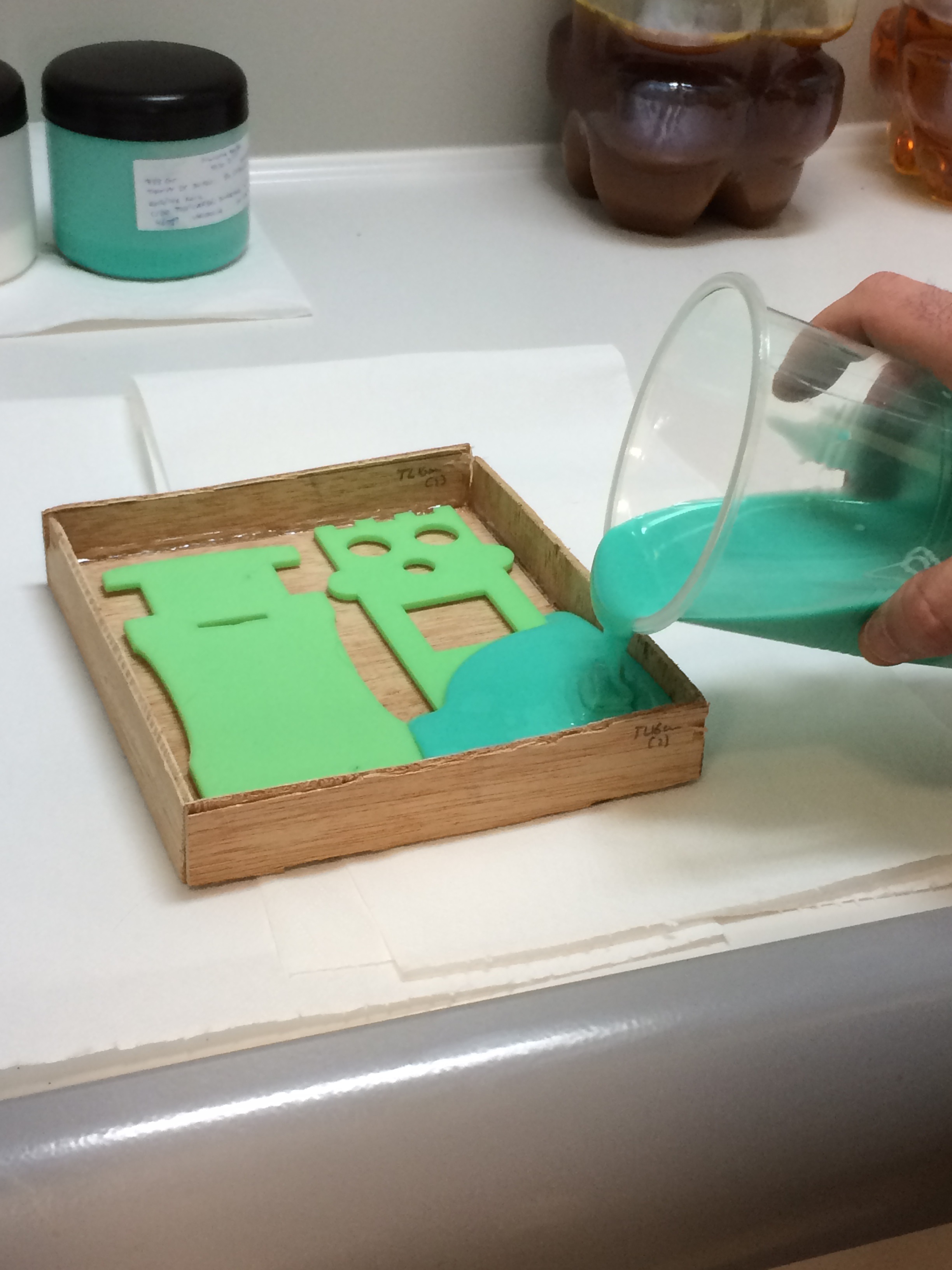}
	\caption{Fabricación de un molde de silicona en el colegio CE Marni (Valencia), 2017}\label{fig:molde_silicona}
\end{figure}

La fabricación de moldes de silicona y creación de copias mediante resinas que puede realizarse en el aula de forma complementaria. Esto puede ser una actividad formativa muy enriquecedora para el alumno que mejorará su motivación. La ventaja principal es que los niños y jóvenes pueden directamente fabricar su propio robot y es una actividad que estoy convencido que les encantará. Quizás en casa, esta actividad carezca de sentido si el objetivo final es fabricar el robot, pero desde luego, si tenéis hijos es una actividad que les encantará (por propia experiencia). En la Figura \ref{fig:molde_silicona} se muestra una foto de la actividad que realizaron en el colegio CE Marni (Valencia) con la fabricación del robot con moldes de forma pionera.

La fabricación de moldes no está exenta de ciertas consideraciones, ya que antes de fabricar un molde debéis pensar en el proceso de desmoldeo, incluso con moldes flexibles como los moldes de silicona. A modo de ejemplo, en nuestros primeros diseños del robot DYOR, incluíamos orificios compatibles con Lego Technic para así poder extender las capacidades del robot, fomentar la creatividad, etc... El problema es que los orificios causaban bastantes problemas de desmoldeo y por este motivo decidimos quitarlos de la propuesta realizada. Llegamos a la conclusión de que era más fácil pegar una barra de Lego Technic allá donde se requiriese.

Con respecto a qué resina utilizar, yo recomiendo dos particularmente: acrílicas y poliuretano, ya que son las que mejor resultado me han proporcionado a un precio muy asumible.

Las resinas acrílicas son muy manejables en aula y principalmente su ventaja es que no son tóxicas, aunque lentas en el proceso de polimerización. Por tanto, si vais a plantear varios robots con estas resinas, deberéis de tener en consideración este hecho, ya que tardaréis algunos días en fabricarlos todos. 

Si por lo contrario, preferís utilizar resinas de poliuretano, debéis de tener en consideración las medidas de precaución indicadas en el propio producto. El tiempo de polimerización es bastante rápido con lo que tenéis poco margen de maniobra, pero su acabado es muy bueno. Se puede fabricar directamente sin fibras y el resultado sigue siendo bueno. En este caso, debéis disponer de un sistema de extracción o ventilación necesaria para evitar intoxicaciones por inhalación. 

Además, para el manejo de cualquiera de las dos resinas se deben utilizar guantes de látex. En el caso de las acrílicas, principalmente para evitar mancharse las manos, mientras que en el caso de las resinas de poliuretano para evitar el contacto con la piel que lleva a ser irritante.

Los costes de los materiales necesarios no tienen porqué ser necesariamente elevados, y por lo general, para el volumen de piezas que se pueden llegar a fabricar, los costes son equiparables. Según nuestra experiencia con el molde de silicona podéis realizar entre 25 y 30 copias utilizando resinas de poliuretano (tóxicas) y estimo que entorno a 100 con resinas acrílicas (no tóxicas), lamentablemente no he llegado a fabricar tantos robots, así que no puedo asegurarlo.

\section{Electrónica y Robótica}

Lo cierto es que la selección de la electrónica dependerá de las funcionalidades que queráis darle al robot, y las soluciones en este punto no son únicas. Yo hago una propuesta que quisiera justificar, pero sois libres de vuestra propia selección, tal y como se menciona en varias ocasiones.

\subsection{Configuración}

Los robots más habituales son los robots con configuración diferencial. Son robots que tienen dos ruedas fijas (de orientación fija), normalmente en la parte trasera y alguna rueda de apoyo que no restrinja la movilidad. Son robots que cuyo movimiento se controla con la velocidad establecida en cada una de las ruedas, de forma que si ambas tienen la misma velocidad, el robot describirá una trayectoria recta; si las ruedas tienen la misma velocidad, pero con sentido contrario, el robot girará sobre sí mismo; y si una de las ruedas tiene una velocidad mayor que la otra, el robot describirá un círculo de un radio determinado. Por sencillez de construcción, esta configuración es sin duda la más recomendable.

\begin{figure}
	\centering
	\subfigure[Configuración diferencial]{
	\includegraphics[width=0.48\columnwidth]{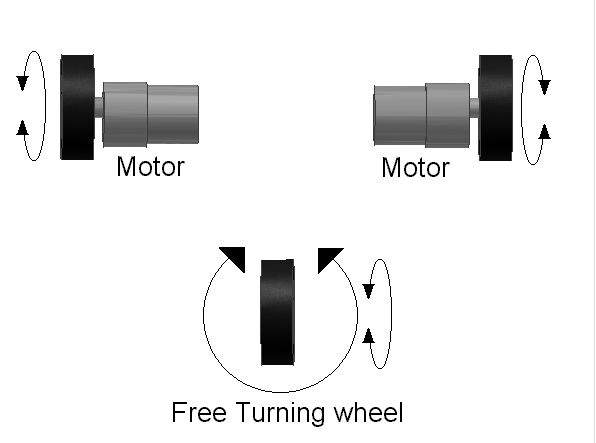}}
    \subfigure[Configuración Ackermann]{
    \includegraphics[width=0.48\columnwidth]{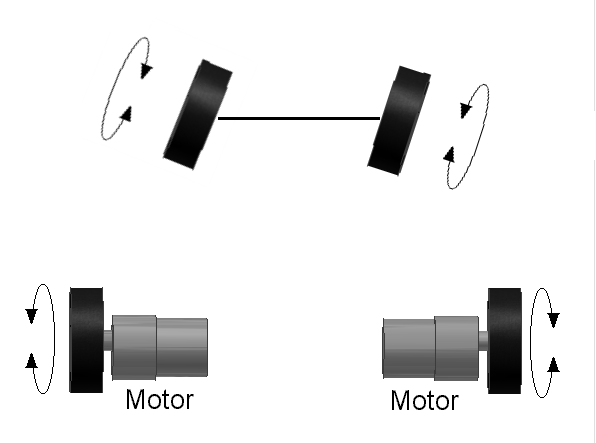}}
    \subfigure[Robot omnidireccional de SeedStudio]{
    	\includegraphics[width=0.75\columnwidth]{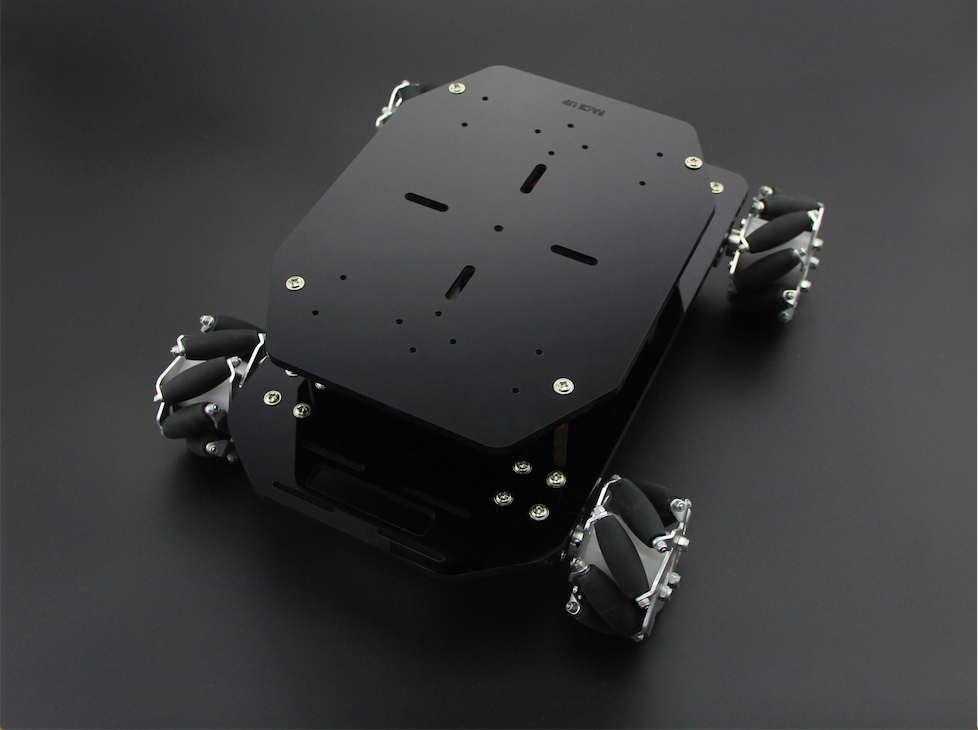}}
	\caption{Configuraciones típicas de los robots móviles. Figuras (a) y (b) creadas por Patrik en wikibooks.}
\end{figure}

Otra configuración muy típica son los robots con configuración tipo triciclo o coche, en la que tienen una rueda orientable, normalmente motorizada, o dos ruedas coordinadas con el mecanismo de Ackermann. Por otro lado una rueda o varias realizan la función de tracción. La principal ventaja de esta configuración es que es maniobrable, quiere decir, que especificando la orientación a la rueda orientable podemos determinar en qué dirección queremos que el robot se mueva. No obstante, en realidad con las transformaciones cinemáticas adecuadas se puede controlar un robot diferencial como si fuera un robot del tipo triciclo y viceversa.

Otra configuración que típicamente podréis encontrar en los robots es la configuración omnidireccional. Este robot tiene el principal atractivo que es capaz de moverse instantáneamente en cualquier dirección, tanto hacia delante o atrás como hacia lo lados. También es capaz de girar sobre sí mismo. La clave está en el tipo de ruedas que utiliza, conocidas como ruedas Suecas, omnidireccionales o \emph{meccanum}. Tened en consideración que necesitaréis controlar cuatro motores, con lo que el precio de estos robots suele ser mayor que el resto.

También, dentro del sistema de locomoción, algunos robots incorporan cadenas o gomas para trasmitir el movimiento a otras ruedas. Son soluciones más que adecuadas en función del tipo de robot que queráis fabricar. Por lo general un robot con cadenas tendrá mejor agarre que un robot con ruedas convencionales.

\subsection{Arduino, shields y comunicaciones}

Bueno, a estas alturas ya debéis de saber que trabajaremos con Arduino y por tanto, lo primero que deberíamos plantearnos es ¿qué Arduino escojo? Lo cierto es que esto depende de vuestro propio criterio, ya que hay varias alternativas factibles.

Mi sugerencia es utilizar Arduino Nano v3.0 (Figura \ref{fig:nano})y el shield o placa de expansión de entradas y salidas (Figura \ref{fig:nano_expansion}), frente a los kits de iniciación de la gran mayoría de los modelos de Arduino. Los motivos son varios:
\begin{itemize}
	\item Precio
	\item Accesibilidad de los pines de conexión
	\item Funcionamente es 100\% compatible con Arduino Uno
\end{itemize}

La única desventaja es que la gran mayoría de shields de Arduino están diseñados para ser compatibles con Arduino Uno (aunque existen placas de adaptación) y por tanto no podríais utilizarlos directamente. No obstante, en la propuesta que yo hago, esto no va a ser necesario y por tanto no representa un problema.

Además, para dotar al robot de comunicación inalámbrica, se plantea el uso de un asequible dispositivo bluetooth, el módulo HC-06 (Figura \ref{fig:bluetooth}). Existen varias variantes de Arduino que disponen ya de comunicaciones Wi-Fi, Zig-bee y bluetooth integradas, pero por lo general son mucho más caras que el conjunto Arduino Nano v3.0, placa expansión E/S y módulo bluetooth HC-06.

Uno de los principales inconvenientes, aunque todo es relativo, de la solución que planteo es que para programar el robot debemos hacerlo por cable USB. Si quisiéramos programar nuestro robot a través de bluetooth, entonces recomiendo comprar el módulo HC-05, similar al HC-06, pero con capacidad de ejercer de \emph{maestro} de las comunicaciones. En definitiva, con una electrónica sencilla (que requiere la fabricación de una placa o de un circuito electrónico adicional con protoboard) se puede resetear a Arduino para que entre en el modo programación. Esto está explicado en un de los vídeos del curso, aunque está fuera de la propuesta del robot realizado.

\subsection{Motores}
Los motores son una de las partes más importantes de un robot y su selección condicionará parte de la electrónica adicional que necesitemos. Personalmente, con objeto de abaratar costes, soy partidario de trabajar con micro servos, ya que incorporan tanto la electrónica como la mecánica necesaria en un encapsulado muy compacto (y relativamente robusto). La principal ventaja es que con una sola señal, podéis controlar tanto posición o velocidad (según el tipo de servo).

De hecho los micro servos de posición estándar SG90 (Figura \ref{fig:sg90}) o similares son muy baratos y os permitirán controlar la posición de determinados ejes (de 0º a 180º). En mi propuesta los utilizo para controlar las pinzas del robot. Por otro lado, los micro servos de rotación continua FS90R (Figura \ref{fig:fs90r}) permite controlar la velocidad de un eje. Eléctricamente y mecánicamente ambos servos son distintos, aunque en apariencia son muy similares. Además, Arduino Nano v3.0 es capaz de alimentarlos directamente sin necesidad de fuentes de alimentación externas. Todo esto realmente representa una ventaja importante desde el punto de vista de sencillez y comodidad en el diseño electrónico, ya que no requiere de componentes adicionales para el control, ni soldaduras ni de circuitos de potencia y por tanto, considero que tanto en el aula, como en casa, donde quizás no dispongáis de los medios adecuados, es la solución más idónea.

Una alternativa razonable por ser algo más baratos, es utilizar los conocidos micromotores de Pololu con engranajes. Estamos probablemente hablando de un ahorro de unos 4€ o 5€, pero no mucho más. A mi juicio, esta solución tiene dos desventajas primordiales y es que consume dos pines de Arduino para poder controlar tanto la velocidad como el sentido de giro y además requiere de soldadura, con lo que quizás, y todo esto depende del colegio o la persona particular en casa, no sea la actividad más idónea, aunque entiendo que no lo descartéis.

\subsection{Sistema de sensorización}

Los robots hoy en día incorporan todo tipo de sensores. Desafortunadamente, la gran mayoría de ellos se salen del presupuesto al respecto de materiales de bajo coste o implican el uso de otras electrónicas de mayor capacidad de procesamiento que también incrementan el coste.

Personalmente opino que son necesarios en los robots sensores con capacidad de medir obstáculos del entorno, también conocidos como sensores de rango. De entre todos los sensores de rango, el más idóneo para el tipo de robot que vais a diseñar es el sensor de ultrasonido SR-HC04 (Figura \ref{fig:sr-hc04}), por su bajo coste y por los rangos de medición que son capaces de alcanzar. Los ultrasonidos emiten ondas ultrasónicas a una determinada frecuencia que se propagan por el aire y que tras rebotar sobre los objetos, estas ondas son devueltas y \emph{escuchadas} por el propio sensor. La señal dista mucho de ser perfecta, ya que está sujeta a todo tipo de fuentes de error que afectan a la medición. Esto se nota particularmente en un aula, cuando tenemos muchos sensores del mismo tipo emitiendo a la misma frecuencia y al mismo tiempo. Aún así, considero que es la mejor opción disponible.

Los sensores de infrarrojos pueden utilizarse o bien para medir distancias o bien para detectar luz. Son idóneos, también por su bajo coste para implementar aplicaciones de seguimiento de línea. El módulo TCRT5000 (Figura \ref{fig:tcrt5000}) permite obtener la cantidad de luz recibida en dos señales, una analógica y otra digital. De forma similar, los módulos sigue-luz permiten proporcionar la cantidad de luz recibida por dos sensores fotosensibles y que pueden utilizarse en aplicaciones en las que queremos que el robot se comporte como si fuera un bicho.

Los sensores de color como el TCS3200 son, por lo general, células fotosensibles (como un píxel de una cámara) que proporcionan sensibilidad diferentes para colores diferentes. En realidad, suelen tener un LED (o varios LEDs) que emite una luz en un color determinado y si el objeto es de ese mismo color, tenemos más sensibilidad a que la luz sea devuelta que si fuera de otro color. Son idóneos para detectar zonas de colores en las que queremos que nuestro robot se muevan.

Los encoders son sensores que os permitirán medir la cantidad de pasos o vueltas que dan vuestras ruedas. Tienen una importancia vital en la gran mayoría de sistemas robóticos, ya que nos pueden ayudar a estimar la posición. Sin embargo no están exentos de errores que hacen que al final la posición tenga una deriva. Por tanto, salvo que los combinemos con sensores de rango y lo hagamos correctamente utilizando técnicas de fusión sensorial bayesianas para construir mapas y localizar la robot. Por todos estos motivos, no estoy personalmente convencido de su utilidad para proyectos de bajo coste como el que planteo. En este sentido, aplicaciones en las que el robot realiza comportamiento semi-elaborados utilizando sensores como ultrasonidos e infrarrojos acabarán siendo, por lo general más interesantes que el hecho de seguir una trayectoria determinada. Aún así, hay aplicaciones en las que medir la cantidad de desplazamiento realizado por el robot es necesario. En ese caso, es más que probable que necesitemos encoders. El problema principal es que, dependiendo del tipo de sensor, pueden llegar a ser relativamente caros, aunque hay soluciones de muy bajo coste utilizando fotodetectores o compases magnéticos. También hay soluciones en las que el motor y el encoder están integrados, junto con la electrónica de potencia y de lectura del encoder, pero esto encarecerá la solución propuesta. En el caso de que tengamos que detectar flancos de subida y/o bajada de las señales de un encoder, necesitaremos utilizar o bien una electrónica de contador o interrupciones, pero desafortunadamente Arduino Nano v3.0 no es precisamente la electrónica más idónea para esto, ya que tan sólo dispone de dos interrupciones, mientras que por lo general deberíamos necesitar cuatro (dos por cada sensor). 

Los sensores acelerómetros y compases electrónicos se pueden utilizar, aunque por lo general en menor medida. Los primeros porque está sujetos a derivas y offsets que hacen que el proceso de integración que se debe realizar para traducir dicha aceleración no sean válidos para el cálculo de la posición, pero al menos sí, para el cálculo del vector gravedad, utilizado en muchos tipos de robots. Por otro lado, los compases electrónicos (o brújulas electrónicas), que miden el campo magnético, suelen verse fuertemente afectados por las variaciones de campo magnético, ya que este es muy débil en comparación de las ondas electromagnéticas o por encontrarse bajo la influencia de algún metal o motor cercano.

Otros sensores más avanzados como cámaras, GPS, entre otros se salen completamente del planteamiento, ya que requieren por lo general de procesadores más avanzados, con un mayor coste. En ese caso, soy partidario del uso de la cámara de los dispositivos propios dispositivos móviles y que hoy día muchos alumnos ya disponen de ellos.

\subsection{Otras electrónicas}

Tradicionalmente se han utilizado pantallas LCDs con el propósito de mostrar información interna del robot, si bien, creo que para el tipo de robot planteado, es mucho más enriquecedor el uso de una matriz de LEDs 8x8 (Figura \ref{fig:matriz-leds}), ya que permite generar todo tipo de dibujos y expresiones en el robot, además de poder ser utilizada para depuración de nuestro programa.

Por otro lado, el uso de zumbadores de sonido como el KY-006 (Figura \ref{fig:zumbador}) son también muy recomendables, ya que os permitirán generar melodías musicales que la primera vez hacen mucha gracia y cuando lleváis 100 veces escuchando la misma melodía quizás no os haga tanta gracia...

\subsection{Alimentación}

La alimentación de un robot es también un aspecto muy importante a considerar. Tened en consideración que vuestro robot está, por lo general, combinando elementos o dispositivos que no requieren grandes cantidades de corriente y otros que sí que demandan mayores consumos de corriente como son los motores o incluso el ultrasonidos (sobretodo demanda un pico de corriente elevado). Los pines de salida de los microcontroladores soportan muy poca corriente y además hay que considerar que el consumo total de todos los pines, por lo general, no supera los 40mA. Por tanto, la corriente debe de venir del sistema de alimentación.

En la solución que propongo el suministro de corriente se simplifica bastante. Arduino Nano v3.0 se puede alimentar o a través del pin Vin, típicamente a 9V, pasando a través de un regulador de tensión o a través del propio puerto USB a 5V. La fila de pines de alimentación de la placa de expansión de E/S están conectados a la salida de +5V de Arduino que soporta hasta unos 800mA de corriente como máximo (de forma aproximada). Afortunadamente, esto está dentro de los requerimientos de consumo habituales de la electrónica propuesta y por tanto podremos conectar los motores directamente alimentados por el pin de +5V de Arduino e indirectamente por la batería USB. Cuando el robot se conecta a un puerto USB de un PC con 500mA de corriente de salida el robot sigue funcionando perfectamente.

Otras alternativas de alimentación es el uso de baterías LI-PO, con resultados similares al uso de baterías USB, pero con menores requerimientos de espacio. Personalmente desaconsejo el uso de pilas de +9V o pilas recargables de 1.2V, suelen durar muy poco y el robot presenta problemas de caídas de tensión que hace que se reinicie con frecuencia. 

\subsection{Propuesta electrónica}

\begin{figure}
	\centering
	\subfigure[Arduino Nano v3.0]{\includegraphics[width=0.32\columnwidth]{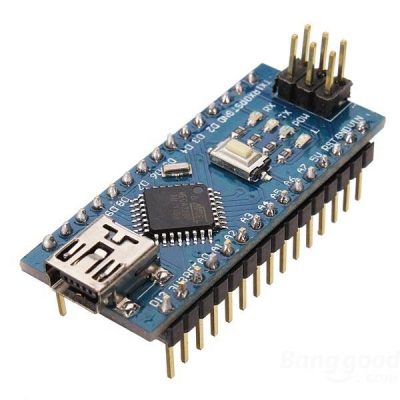}\label{fig:nano}}
	\subfigure[Placa expansión de E/S de Arduino Nano]{\includegraphics[width=0.32\columnwidth]{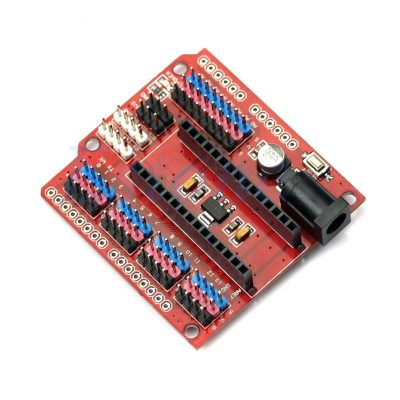}\label{fig:nano_expansion}}
	\subfigure[Módulo Bluetooth HC-06]{\includegraphics[width=0.32\columnwidth]{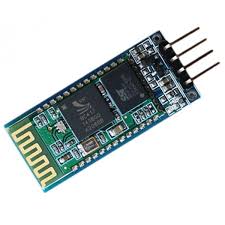}\label{fig:bluetooth}}
	\subfigure[Modulo seguilíneas]{\includegraphics[width=0.32\columnwidth]{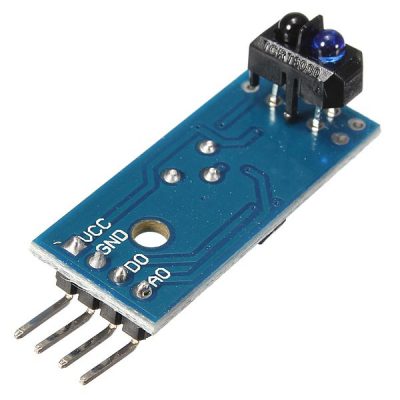}\label{fig:tcrt5000}}
	\subfigure[Sensor de ultrasonidos]{\includegraphics[width=0.32\columnwidth]{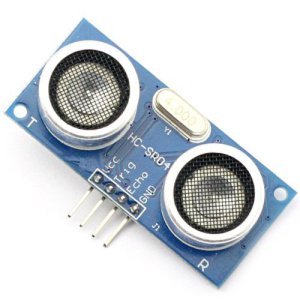}\label{fig:sr-hc04}}
	\subfigure[Matriz LEDs]{\includegraphics[width=0.2\columnwidth]{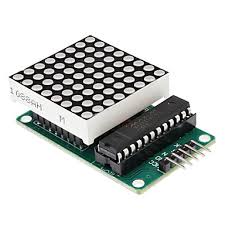}\label{fig:matriz-leds}}
	\subfigure[Microservos de posición]{\includegraphics[width=0.32\columnwidth]{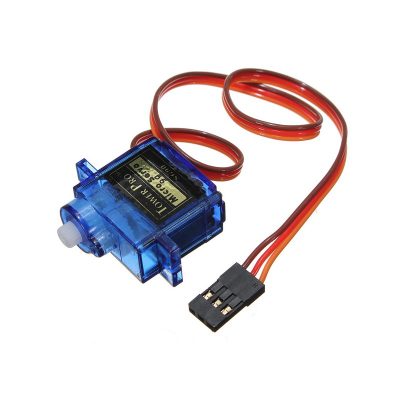}\label{fig:sg90}}
	\subfigure[Microservos de rotación contínua]{\includegraphics[width=0.32\columnwidth]{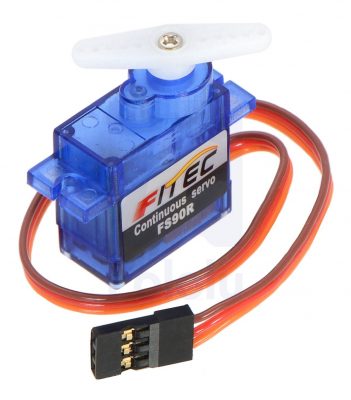}\label{fig:fs90r}}
	\subfigure[Zumbador de sonido]{\includegraphics[width=0.32\columnwidth]{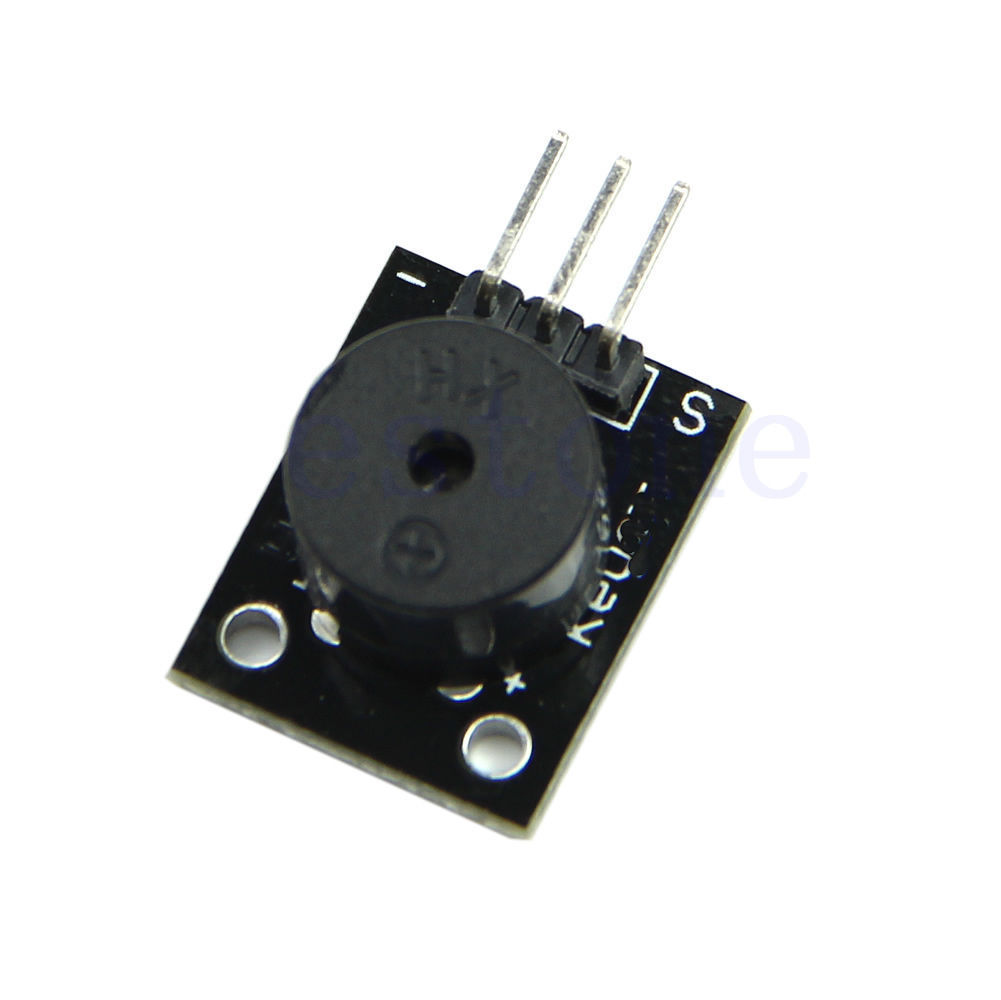}\label{fig:zumbador}}
	\caption{Electrónica principal propuesta para el robot DYOR}\label{fig:electronica_dyor}
\end{figure}

A continuación se detalla la propuesta electrónica (ver Figura \ref{fig:electronica_dyor}). Tiene un coste aproximado de unos 65€:
\begin{itemize}
	\item Arduino Nano v3.0 con cable USB
	\item Placa de expansión de E/S
	\item 2xMicroservos FS90R (de rotación continua)
	\item 2xMicroservos SG90 (de posición)
	\item Módulo de seguimiento de línea con sensor infrarrojo TCRT5000
	\item Sensor de ultrasonidos para medir distancias HC-SR04
	\item Matriz de LEDs 8x8 (max7219)
	\item Módulo bluetooth HC-06
	\item Zumbador de sonido KY-006
	\item Batería de alimentación (powerbank) USB
	\item Rueda loca
	\item Cables Hembra-Hembra DuPont
\end{itemize}

\begin{figure}
	\centering
	\includegraphics[width=\columnwidth]{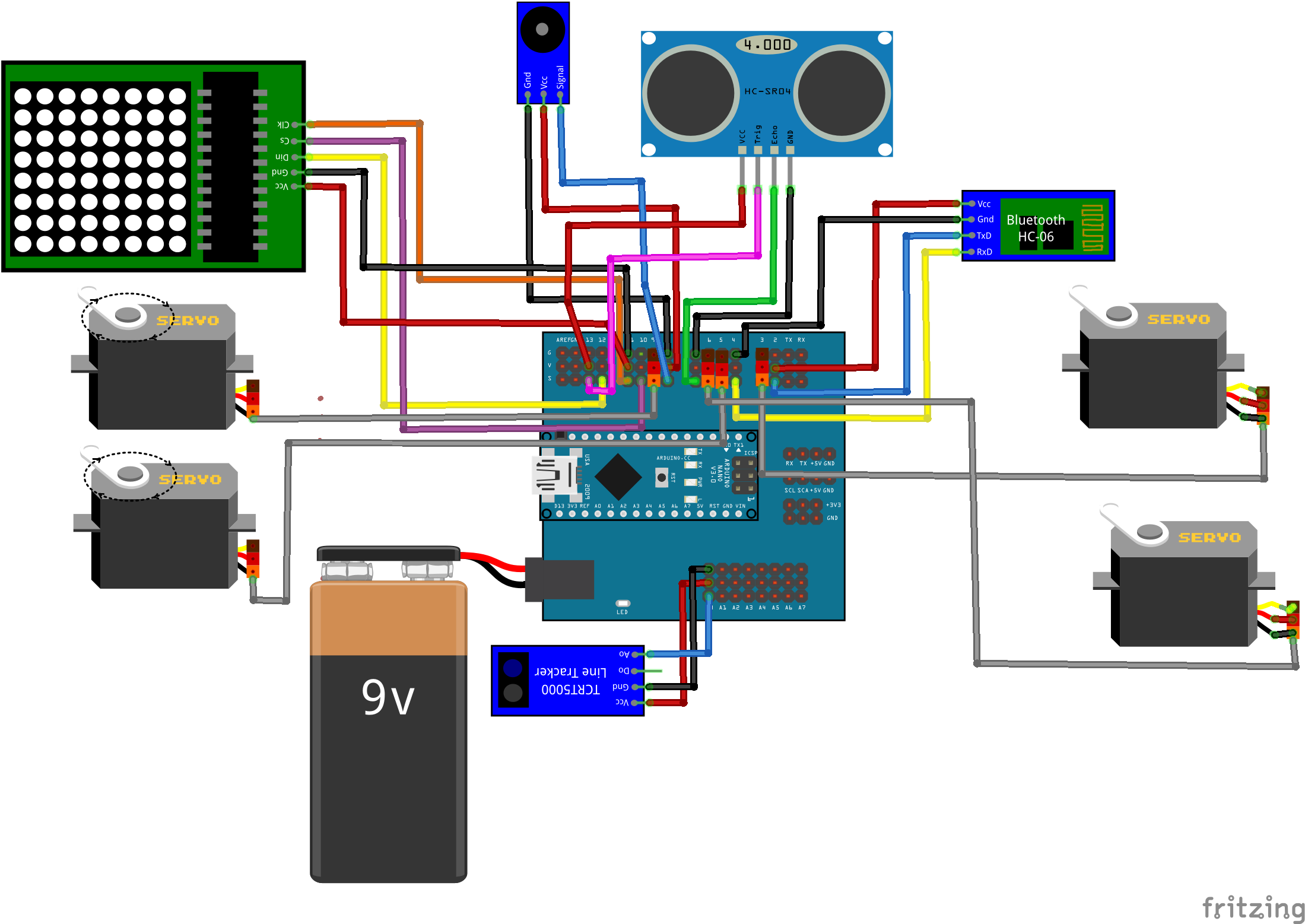}
	\caption{Esquema electrónico propuesto}\label{fig:esquema_electronico}
\end{figure}

En la Figura \ref{fig:esquema_electronico} se muestra el esquema electrónico asociado a la propuesta (en la figura aparece una pila de 9V que en realidad ha sido reemplazado por la batería de alimentación USB). Podéis observar que no es necesario realizar ningún tipo de soldadura con la propuesta realizada, tan sólo debemos conectar cables entre los diferentes dispositivos y la placa de expansión de E/S de Arduino Nano. El esquema ha sido realizado con Fritzing, un programa para el diseño electrónica de fácil manejo que se explica en el curso.

\section{Programación}

Para la programación del robot, la apuesta es utilizar lenguajes de programación por bloques. Son muy intuitivos de utilizar y la curva de aprendizaje es muy rápida.

\subsection{Facilino}

\begin{figure}
	\subfigure[Código Facilino para el control de un servo]{\includegraphics[width=0.35\columnwidth]{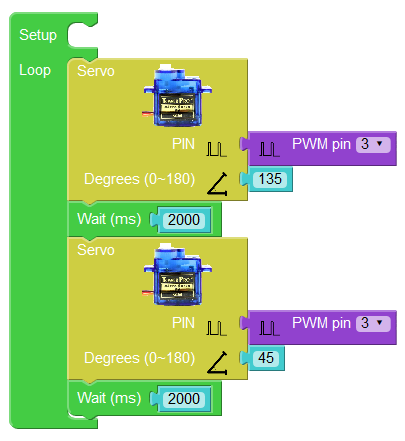}}
	\subfigure[Código Facilino para generar una melodía]{\includegraphics[width=0.55\columnwidth]{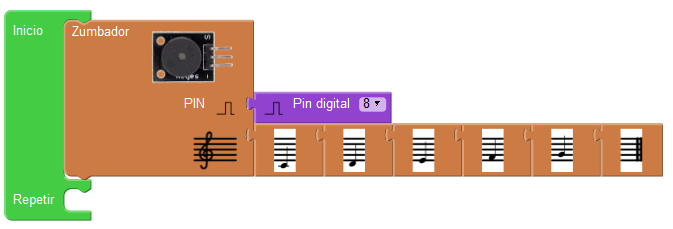}}\\
	\subfigure[Código Facilino para mostrar dibujos en la matriz de LEDs]{\includegraphics[width=0.45\columnwidth]{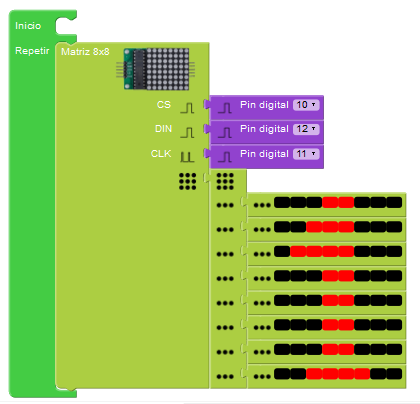}}
	\subfigure[Código Facilino para evitar colisiones]{\includegraphics[width=0.45\columnwidth]{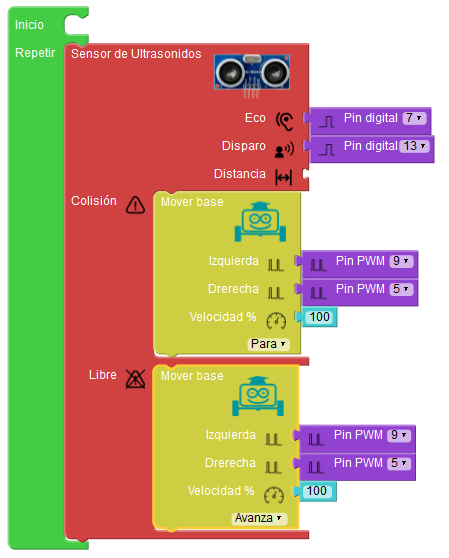}}
	\caption{Algunos ejemplos de código Facilino}
\end{figure}

Facilino es un programa de programación por bloques de Arduino. Considero que Facilino cubre todas las funcionalidades que este curso requiere y por eso planteo su uso. Facilino dispone de una versión gratuita con funcionalidades de bajo nivel para el acceso a las señales de Arduino, así como el control de flujo del programa. En ese sentido es similar a Bitbloq con la principal diferencia de que soporta una gran variedad de electrónicas de Arduino estándar, más asequibles que por ejemplo la ZumBoard. La versión Pro de Facilino requiere de una licencia a un precio razonablemente asequible que os proporcionará funcionalidades que utilizaremos a lo largo del curso, porque simplifican mucho el código a implementar. Además, de hecho, comprando los kits de electrónica del robot DYOR en realidad esta licencia es gratuita.

Facilino proporciona funcionalidades de alto nivel que simplifican los procesos de comunicación por bluetooth mediante la recepción de comandos y el envío y recepción de telegramas. También proporciona funcionalidades para el manejo de la matriz de LEDs, generación de melodías musicales de forma intuitiva, implementación de sencillas funcionalidades para seguir líneas o evitar obstáculos y funcionalidades para mover los motores de robot (en configuración diferencial) de forma coordinada (a partir de su velocidad lineal y angular).

Desaconsejo el uso de S4A en este proyecto porque en realidad el código Scratch no se ejecuta en el procesador de Arduino, si no que es el PC el que comunica con Arduino a través de un firmware y por tanto requiere de una conexión permanente a través del cable USB. Como consecuencia, el robot no sería autónomo.

Si no pensáis trabajar por bloques, entonces recomiendo utilizar directamente Arduino IDE. El problema, es como ya he comentado, es que es menos intuitivo de utilizar para personas que se inician en el mundo de la electrónica y los niños.

\subsection{App Inventor2}

App Inventor2 es un programa idóneo para la programación por bloques de Apps en Android. Dispone de un gran conjunto de funcionalidades como el uso componentes habituales en las interfaces de usuario; elementos para la disposición de componentes a través de contenedores; componentes para acceder a los sensores del dispositivo móvil (acelerómetro, reloj, GPS, etc); acceso a redes sociales; conectividad web y bluetooth y recientemente han abierto las posibilidades al permitir el uso de extensiones.

\begin{figure}
	\centering
	\includegraphics[width=0.7\columnwidth]{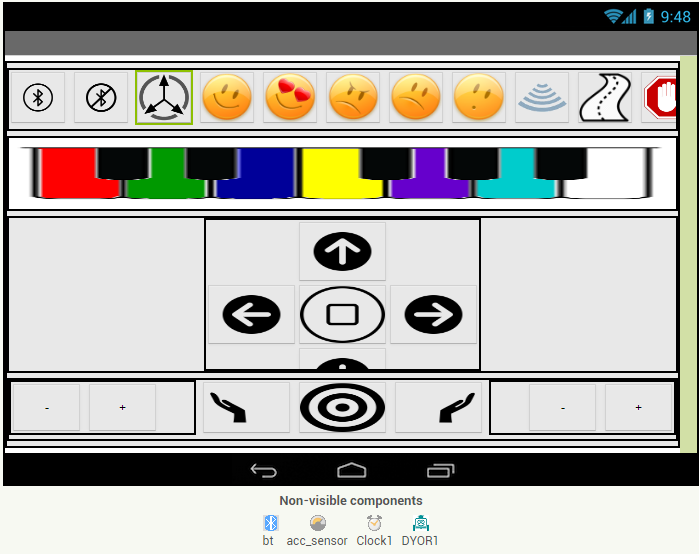}
	\caption{Aplicación propuesta para el control remoto del robot DYOR}\label{fig:control_remoto}
\end{figure}

En el curso se explica el uso de los componentes principales de App Inventor2, al mismo tiempo que se explican y proporcionan varios ejemplos de aplicaciones sencillas. Además, se explica cómo desarrollar una aplicación para el control remoto del robot, tal y como se muestra en la Figura \ref{fig:control_remoto}.

\section{Posibles aplicaciones a implementar}

\subsection{Control remoto}
El control remoto se realiza con el dispositivo móvil, utilizando su acelerómetro o a través de una botonera. También se pueden controlar remotamente funcionalidades diversas como la emisión de notas musicales o expresiones en la matriz de LEDs. Esto es lo que yo propongo, pero lo cierto es que existen muchas más funcionalidades que podrían llegar a implementarse.

\subsection{Autónomas}

\subsubsection{Seguimiento de líneas}

En el curso se explica cómo implementar una aplicación de seguimiento de líneas utilizando el sensor de infrarrojos. El sensor permite claramente discriminar entre un fondo blanco y un fondo negro (típicamente la línea). Hay personas que utilizan dos sensores para hacer un seguimiento de líneas, pero en realidad con un solo sensor que siga en realidad el borde de la línea sería suficiente. La aplicación propuesta es en realidad bastante sencilla ya plantea el control de la velocidad angular del robot de forma proporcional a un valor de luz de referencia (típicamente el valor intermedio entre el valor de lectura del color blanco y del negro).

\subsubsection{Sigue-luz}
En este caso necesitaréis un módulo sigue-luz (no está dentro de la propuesta de electrónica), pero el comportamiento a implementar es muy sencillo, ya que la velocidad lineal se puede regular en función de la cantidad de luz recibida y la velocidad angular se puede regular en función de la diferencia de luz recibida por los sensores.

\subsubsection{Sumo}

En este tipo de aplicaciones, el robot debe ser capaz de expulsar a otro robot fuera del entorno (ring o arena) marcado típicamente por una línea negra. El problema es que el rival tiene el mismo objetivo y por tanto se trata de ser capaz de evitar las posibles embestidas y embestir lo más rápido posible tras detectar la posición del rival. Hay estrategias que consisten en hacer movimientos rápidos hasta detectar el final del ring , girarse rápidamente de forma pseudo-aleatoria y continuar así hasta detectar al rival. En ese momento se debe apuntar directamente a por él y tratar de expulsarlo. Desafortunadamente, la propuesta del robot DYOR no es precisamente la más idónea para este propósito, lo cual no quiere decir que no se pueda implementar la aplicación. Esto es porque normalmente los robots que participan en este tipo de pruebas suelen ser robots robustos y con mucho par de motor.

\subsubsection{Recoge piezas}
El objetivo de esta aplicación es recoger un conjunto de piezas en el menor tiempo posible. El robot debe tratar primer de localizar una pieza y en el momento en que la detecta debe de tratar de empujarla hasta el final del ring o arena (el sensor de infrarrojos se puede utilizar con éste propósito).

\subsubsection{Evitación de obstáculos}
Existen muchas formas de evitar obstáculos. Lógicamente una de las claves para poder hacerlo correctamente es poder disponer de la información necesaria de donde están los obstáculos. El problema es que con la electrónica propuesta estamos realmente muy limitados y sólo detectaríamos obstáculos en la parte frontal. Normalmente cuando detectamos un obstáculo, lo haremos a partir de una cierta distancia umbral que consideraremos peligrosa. En ese momento debemos asegurarnos que el robot para, gira en algún sentido y se vuelve a mover sólo cuando el obstáculo ya no está presente.
Hay robots que incorporan sensores también en los laterales, para poder ayudar al proceso de decisión de donde girar y hasta cuanto girar. Una técnica rudimentaria, pero bastante efectiva es simplemente girar un tiempo aleatorio. En caso de que el obstáculo todavía estuviera delante, porque el tiempo de giro no ha sido suficiente, siempre podremos volver a girar (preferiblemente recordando el último sentido de giro).

\subsubsection{Expresiones y melodías}
Lo cierto es que la matriz de LEDs propuesta para el robot DYOR y el zumbador dan mucho juego para enriquecer las aplicaciones que desarrolléis, porque se pueden programar fácilmente gracias a las instrucciones proporcionadas por Facilino.

\bibliography{biblio}{}
\bibliographystyle{plain}

\end{document}